\documentclass{article}
\usepackage{ijcai25}

\usepackage{times}
\usepackage{soul}
\usepackage{url}
\usepackage[utf8]{inputenc}
\usepackage[small]{caption}
\usepackage{graphicx}
\usepackage{amsmath}
\usepackage{amsthm}
\usepackage{booktabs}
\usepackage{algorithm}
\usepackage{algorithmic}
\usepackage[switch]{lineno}

\usepackage{amsfonts}
\usepackage{array}
\usepackage{threeparttable}
\usepackage{multirow}
\usepackage{makecell}
\usepackage{booktabs} 
\usepackage{colortbl} 
\usepackage{bm}
\usepackage{url}
\title{Dual-Balancing for Physics-Informed Neural Networks}

\author{
Chenhong Zhou$^1$
\and
Jie Chen$^1$\and
Zaifeng Yang$^2$\And
Ching Eng Png$^2$\\
\affiliations
$^1$Department of Computer Science, Hong Kong Baptist University, Hong Kong SAR, China\\
$^2$Institute of High Performance Computing (IHPC), Agency for Science, Technology and Research (A*STAR), Singapore\\
\emails
\{cschzhou, chenjie\}@comp.hkbu.edu.hk,
\{yang\_zaifeng, pngce\}@ihpc.a-star.edu.sg
}

\begin{document}

\maketitle
\begin{abstract}
Physics-informed neural networks (PINNs) have emerged as a new learning paradigm for solving partial differential equations (PDEs) by enforcing the constraints of physical equations, boundary conditions (BCs), and initial conditions (ICs) into the loss function. Despite their successes, vanilla PINNs still suffer from poor accuracy and slow convergence due to the intractable multi-objective optimization issue. In this paper, we propose a novel Dual-Balanced PINN (DB-PINN), which dynamically adjusts loss weights by integrating inter-balancing and intra-balancing to alleviate two imbalance issues in PINNs.  Inter-balancing aims to mitigate the \textbf{\textit{gradient imbalance}} between PDE residual loss and condition-fitting losses by determining an aggregated weight that offsets their gradient distribution discrepancies. Intra-balancing acts on condition-fitting losses to tackle the \textbf{\textit{imbalance in fitting difficulty}} across diverse conditions. By evaluating the fitting difficulty based on the loss records, intra-balancing can allocate the aggregated weight proportionally to each condition loss according to its fitting difficulty level. We further introduce a robust weight update strategy to prevent abrupt spikes and arithmetic overflow in instantaneous weight values caused by large loss variances, enabling smooth weight updating and stable training. Extensive experiments demonstrate that DB-PINN achieves significantly superior performance than those popular gradient-based weighting methods in terms of convergence speed and prediction accuracy. Our code and supplementary material are available at \url{https://github.com/chenhong-zhou/DualBalanced-PINNs}.


\end{abstract}

\section{Introduction}
Accurately and efficiently solving the partial differential equations (PDEs) governing physical systems provides significant advantages for understanding their spatiotemporal evolution~\cite{karniadakis2021physics,wandel2022spline,wei2024pdenneval}. In recent years, physics-informed neural networks (PINNs) \cite{raissi2019physics} have emerged as a new learning paradigm for solving PDEs. PINNs enforce the network to satisfy the constraints given by the PDEs, initial conditions (ICs), and boundary conditions (BCs) via minimizing a composite loss function. The loss in PINNs typically comprises a PDE residual loss evaluated on a sample of scattered interior points (referred to as collocation points), and loss terms for initial and boundary points. In addition, a prominent advantage of PINNs is seamless integration of observational data with physical laws. This integration bridges the gap between data-driven learning and physics-driven learning, thereby offering a powerful framework for a wide range of scientific applications \cite{karniadakis2021physics,shi2021physics,meng2022physics,ashraf2024physics,li2024causality,chu2024structure}. However, PINNs sometimes have been found to perform poorly and even fail to converge, especially when solving stiff PDEs whose solutions exhibit fast-paced spatial-temporal transitions \cite{krishnapriyan2021characterizing,kim2021dpm,wang2022and}. Previous studies have revealed that a leading cause of failure is attributed to the imbalanced multi-objective loss optimization \cite{wang2022and,wang2021understanding}. Typically, the PDE residual loss dominates the total training process and converges faster, while certain condition-fitting losses (collectively referring to the loss terms of IC, BCs, observations, etc.) suffer from vanishing back-propagated gradients \cite{wang2022and,wang2021understanding}. The trained model is strongly biased toward yielding a solution with a small PDE residual error, ignoring the accurate fitting of initial and boundary conditions. This is why PINNs often fail to learn the correct solution.

Numerous studies have made efforts to balance the interplay between different terms in the composite loss function by tuning weights for loss components \cite{groenendijk2021multi,dashtbayaz2024physics,song2024loss}. Wang \emph{et al.} \cite{wang2021understanding} propose to dynamically tune weights based on the mean magnitude of the backpropagated gradients of the loss function. This work serves as a foundational contribution, inspiring many follow-up dynamic weighting methods based on different gradient statistics, e.g., standard deviation \cite{maddu2022inverse}, kurtosis \cite{vemuri2023gradient}, and Euclidean norms \cite{deguchi2023dynamic}. These gradient statistics-based weighting methods can alleviate the dominance of PDE residual loss by emphasizing the contributions of condition-fitting losses. Specifically, they consider the PDE residual loss as a baseline and assign the weights to condition-fitting losses by relating each condition-fitting loss's gradients to the PDE residual loss's gradients, in order to mitigate the gradient imbalance. However, we notice that these methods are merely based on the pairwise relationship between the PDE residual and each condition-fitting loss. They ignore the intra-relations within condition losses and do not consider the difficulty imbalance for fitting different conditions. This imbalance in condition-fitting difficulty leads to consistently excessive emphasis on easier conditions, thereby neglecting and slowing progress on fitting difficult conditions. Consequently, the model converges slowly and is easily stuck in a local minimum.

To address these issues, we propose a novel Dual-Balanced PINN (DB-PINN), which dynamically adjusts the weights to balance the training process of PINNs. Our proposed DB-PINN integrates \textbf{\textit{inter- and intra-balancing}} to simultaneously alleviate the imbalance in gradient flow between the governing PDE and enforced conditions at the gradient level as well as the imbalance in the condition-fitting difficulty at the loss-scale level. Specifically, \textbf{\textit{inter-balancing}} characterizes the relationship between PDE residual loss and all condition-fitting losses based on the gradient statistics of backpropagation to obtain an aggregated weight. On the other hand, \textbf{\textit{intra-balancing}} aims to evaluate the fitting difficulty of each condition based on its training losses and then allocate the aggregated weight proportionally to these condition-fitting losses. Hence, intra-balancing prevents unnecessary emphasis on fitting easier conditions and allocates larger weights to difficult conditions to accelerate their convergence. Ultimately, the model balances the contributions of PDE residual and condition constraints and also balances the convergence rates among diverse conditions to ensure that their losses converge to an identical level.

Moreover, we observe that the learned weights at each epoch/batch during training are influenced by the stochastic nature of gradient descent updates, resulting in large variance. Instantaneous weight values could suddenly spike, posing a risk of arithmetic overflow. To address this issue, we propose a robust and smooth weight update strategy using Welford's online algorithm \cite{welford1962note}. The proposed update strategy effectively resists large training variances and significantly reduces sharp fluctuations in weight values for smooth weight updating and stable training.

Extensive experiments are performed on several PDE benchmarks to validate the effectiveness of the proposed method. Experimental results show that DB-PINN outperforms popular gradient statistics-based methods by a significant margin in both convergence rate and prediction accuracy.

\section{Background and Related Work }
\paragraph{Physics-Informed Neural Networks (PINNs)} 
The core idea of PINNs \cite{raissi2019physics} is to learn the approximate solution $\hat{u}_\theta(\mathbf{x}, t)$ by constructing a neural network with parameters $\theta$ for solving the following general PDE: 
\begin{align*}
\mathcal{N}_{\mathbf{x}, t}[u(\mathbf{x}, t)]  &= f(\mathbf{x}, t),    & \mathbf{x}\in \Omega, \ t\in[0, T];     \\
\mathcal{B}_{\mathbf{x}, t}[u(\mathbf{x}, t)] &= g(\mathbf{x}, t),    &  \mathbf{x}\in \partial \Omega, \ t\in [0, T];  \\
u(\mathbf{x}, 0) &= h(\mathbf{x}),   & \mathbf{x}\in \Omega, 
  \label{eq1}
\end{align*}
where $\mathbf{x}$ and $t$ are the spatial vector variable and time coordinates, respectively,  $\Omega$ is a bounded spatial domain with the boundary $\partial \Omega$, $u(\mathbf{x}, t)$ is the latent solution to the PDE, $\mathcal{N}_{\mathbf{x}, t}$ and $\mathcal{B}_{\mathbf{x}, t}$ are spatial-temporal differential operators. $f(\mathbf{x}, t)$, $g(\mathbf{x}, t)$, and $h(\mathbf{x})$ are the forcing function, boundary condition, and initial condition functions, respectively. To enforce the approximate solutions satisfying the PDE structure, a set of collocation points $\displaystyle \left \{ (\mathbf{x}^r_i, t^r_i ) \right \}^{N_r}_{i=1}$ are randomly sampled within the entire spatio-temporal domain ($\Omega \times [0, T]$) to compute the PDE residual loss:
\begin{equation}
\mathcal{L}^r(\theta) = \frac{1}{N_r}\sum_{i=1}^{N_r}|| \mathcal{N}_{\mathbf{x}, t}[\hat{u}_\theta(\mathbf{x}^r_i, t^r_i)] - f(\mathbf{x}^r_i, t^r_i) ||^2.
\label{eq_Lr}
\end{equation}
Accordingly, initial and boundary conditions are enforced into the network training by calculating the mean squared loss on the initial and boundary points, respectively. The overall loss function of PINNs can be written as below:
\begin{equation}
\mathcal{L}^{total}(\theta) = \lambda^r\mathcal{L}^r(\theta) + \lambda^{bc}\mathcal{L}^{bc}(\theta) + \lambda^{ic}\mathcal{L}^{ic}(\theta), 
\label{eq2}
\end{equation}
where $\lambda^r$, $\lambda^{bc}$, $\lambda^{ic}$ are loss weights that control the interplay between different loss terms. Observational data can be introduced into Equation (\ref{eq2}) so that PINNs can be applied for solving inverse problems to estimate unknown PDE parameters. Since the inception of PINNs, a wealth of research has been dedicated to improving the performance of PINNs by adaptively resampling collocation points \cite{wu2023comprehensive,yang2023dmis,daw2023mitigating,zhou2024enhanced}, designing novel activation functions \cite{jagtap2020adaptive,gnanasambandam2023self}, and proposing different loss weighting strategies \cite{liu2021dual,xiang2022self,perez2023adaptive}.

\begin{figure*}[!ht] 
  \centering
  \includegraphics[width=\linewidth]{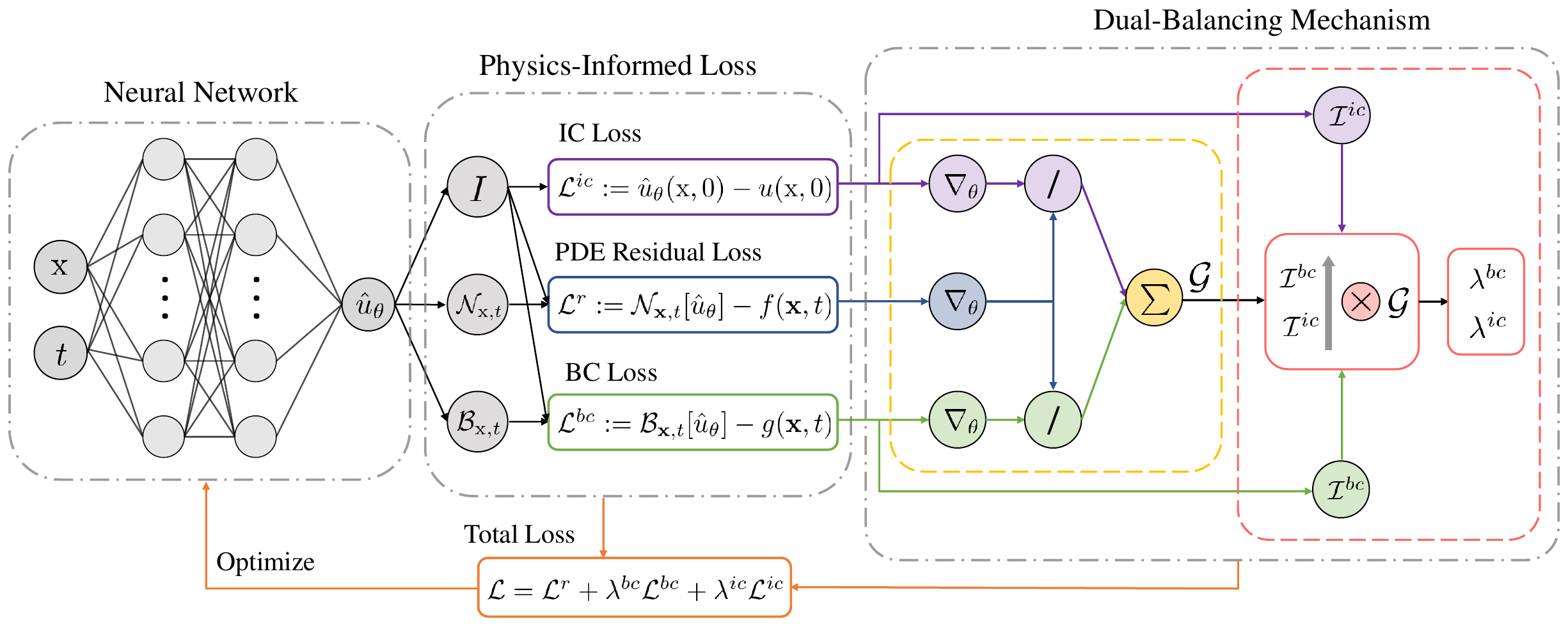}
 \setlength{\abovecaptionskip}{-2pt}
\setlength{\belowcaptionskip}{-5pt}
  \caption{Illustration of the proposed DB-PINN. The flow of different losses is represented via different colors (PDE residual loss: blue; IC loss: purple; BC loss: green). The yellow dashed box signifies inter-balancing, where the gradients ($\nabla_\theta$) of each loss are first calculated, and then gradient distribution discrepancies are evaluated by the ratio ($/$) of $\nabla_\theta \mathcal{L}^r$ to $\nabla_\theta \mathcal{L}^i$ under certain gradient statistical metrics. Total gradient ratio $\mathcal{G}$, also considered as the aggregated weight, is computed by a summation ($\sum$). The pink dashed box signifies intra-balancing, where BC and IC losses are used to calculate their difficulty indexes (i.e., $\mathcal{I}^{bc}$ and $\mathcal{I}^{ic}$). The gray arrow implies that we sort condition losses by fitting difficulty, and finally, the aggregated weight can be proportionally allocated to BC and IC losses to derive their respective weights. 
  }
  \label{fig_PINN}
\end{figure*}

\paragraph{Prior Work on Dynamic Weighting Methods for PINNs}

The loss function in the original formulation of PINNs is composed of an unweighted linear combination of multiple loss components (i.e., all weights set to 1). This approach, referred to as equal weighting (EW), serves as a common baseline, but this baseline often suffers from poor accuracy and slow convergence. Undoubtedly, manually adjusting weights is time-consuming and tedious. Therefore, numerous studies have investigated dynamic weighting methods to adaptively allocate appropriate weights to individual loss components. According to different ways of generating loss weights, existing weighting methods can be mainly categorized into two types: the learning approach and the calculating approach. The learning approach considers the loss weights as learnable parameters and optimizes them together with network parameters, such as the uncertainty weighting (UW) \cite{xiang2022self} and SA-PINN  \cite{mcclenny2020self}. However, the learning approach faces the challenges of increased optimization complexity.

The calculating approach directly computes the weights by analyzing the backpropagation gradient dynamics \cite{wang2021understanding,maddu2022inverse} or through the lens of the neural tangent kernel (NTK) theory \cite{wang2022and}. Specifically, Wang \emph{et al.} \cite{wang2021understanding} observe that the gradients of PDE residual loss generally attain larger values than the gradients of BC loss and IC loss. Based on such findings, they assign appropriate weights to condition-fitting losses to achieve comparable gradient magnitudes and rewrite the total loss as follows: 
\begin{equation} 
\mathcal{L}^{total}(\theta) = \mathcal{L}^r(\theta) + \sum_{i=1}^{M} \lambda^i \mathcal{L}^i(\theta).
\label{eq3}
\end{equation}
The PDE residual loss $ \mathcal{L}^r$ is considered a reference line, and the weight $\lambda^i$ of each condition loss $\mathcal{L}^i$ can be computed based on the ratio of gradient magnitude differences ($\left | \nabla_\theta \mathcal{L}^r \right |/ \left |  \nabla_\theta \mathcal{L}^i \right |$). This work pioneers an effective and impactful weighting method based on backpropagation gradients. Hereafter, several works followed it to utilize and combine different gradients' statistical metrics, such as standard deviation \cite{maddu2022inverse}, kurtosis \cite{vemuri2023gradient}, and Euclidean norms \cite{deguchi2023dynamic}, in order to offer more informative indications of gradient distributions to derive optimal weights. However, this class of gradient statistics-based weighting methods is based on the pairwise relation between $\mathcal{L}^r$ and each $\mathcal{L}^i$, which ignores the intra-relations between $\mathcal{L}^i$ and $\mathcal{L}^j$ ($i\ne j \ne r$, and $i, j \in M$) and does not consider diverse fitting difficulties for different conditions. This class of weighting approach would result in vanishing losses consistently being assigned large weights, thereby biasing the training process towards optimizing easily fitting conditions excessively, while neglecting the slow progress on fitting difficult conditions.

\section{Dual-Balanced PINN (DB-PINN)} 

To address the above issues, we propose DB-PINN by considering the multi-objective optimization in PINNs as two main training goals: (1) balancing the training between the dominant PDE residual loss and condition-fitting losses via inter-balancing, and (2) balancing the training to learn multiple conditions with diverse fitting-difficulty levels via intra-balancing. Furthermore, a robust weight update strategy is introduced for smooth weight updating and stable training.

The proposed DB-PINN integrates inter-balancing and intra-balancing mechanisms to achieve the above two training goals.
Specifically, inter-balancing aims to mitigate the gradient imbalance between the PDE residual loss and all condition-fitting losses based on gradient statistics by measuring the discrepancies between their gradient distributions. Meanwhile, intra-balancing acts between condition-fitting losses to alleviate the imbalance of fitting difficulty by assigning their weights proportionally to the degree of fitting difficulty. The proposed DB-PINN is illustrated in Figure \ref{fig_PINN}.

\subsection{Inter-Balancing Between PDE Residual Loss and Condition-Fitting Losses} 
Previous studies \cite{wang2022and,krishnapriyan2021characterizing,wang2021understanding} have revealed that the PDE residual loss generally exhibits dominance over the other loss terms, and this training imbalance severely hinders the training convergence and predictive accuracy of PINNs. To mitigate this imbalanced training, we first employ gradient statistics-based weighting methods to capture the gradient distribution discrepancies between PDE residual loss and every condition loss. Next, the overall gradient discrepancies can be derived by a summation. Thus, the \textit{total gradient ratio} can be computed and considered as an aggregated weight. Specifically, we formulate the calculation procedure of the aggregated weight by adopting the mean magnitude of gradients (i.e., mean \cite{wang2021understanding}) as a statistic:
\begin{equation}
\mathcal{G}  = \sum_{i=1}^{M} \frac{\mathrm{max}\left \{ \left | \nabla_\theta \mathcal{L}^r   \right |  \right \} }{\overline{\left | \nabla_\theta \lambda^i  \mathcal{L}^i \right | } }, 
\label{totalG}
\end{equation}
where $\nabla_\theta $ denotes the gradient vector of the loss with respect to network parameters $\theta$,  $\left | \cdot \right | $ is the element-wise absolute value of the gradient vector, ${\rm max}$ denotes the maximum magnitude of gradients, and the overbar refers to the mean value of the gradient magnitudes. Equation \eqref{totalG} shows that the total gradient ratio is the sum of the maximum magnitude of the PDE residual loss's gradients divided by the mean magnitude of each weighted condition loss's gradients. Here the discrepancies in gradient distributions are measured by the mean magnitude which can be replaced by other statistical measures, such as standard deviation (std), kurtosis, etc \cite{maddu2022inverse,vemuri2023gradient,deguchi2023dynamic}. As the aggregated weight characterizes the holistic relations between the PDE and conditions, it will be reallocated to each condition-fitting loss via intra-balancing among condition-fitting losses.

\subsection{Intra-Balancing Among Condition-Fitting Losses} 
We further introduce intra-balancing among condition-fitting losses to encourage the model to prioritize fitting difficult conditions and prevent the continual over-allocation of learning resources to easy conditions. First, we define the condition-fitting loss vector $\mathcal{L}_t = [\mathcal{L}^{1},\dots, \mathcal{L}^{M}]^T \in \mathbb{R}_+^M$, where the subscript $t$ is the time step and the superscript denotes the $i$-th condition-fitting loss.
To measure the model's fitting progress for different conditions, we introduce a \textit{difficulty index} $\mathcal{I}$ which is defined as:
\begin{equation} 
\mathcal{I}_t = \frac{\mathcal{L}_t }{\mu_{\mathcal{L}_{t}} }, 
\label{eq5}
\end{equation}
where $\mathcal{L}_t$ is the observed condition-fitting losses at time step $t$, and $\mu_{\mathcal{L}_{t}}$ is the average of the observed losses up to time $t$. The difficulty index $\mathcal{I}^i_t$ for the $i$-th condition reflects its inverse training rate, and thus it quantifies the fitting difficulty for the $i$-th condition. A larger value of $\mathcal{I}^i_t$ corresponds to a slower training rate, which indicates higher difficulty in fitting the $i$-th condition. Correspondingly, this condition loss should be assigned a larger weight. Therefore, we can allocate the aggregated weight proportionally to each condition-fitting loss based on the degree of fitting difficulty: 
\begin{equation} 
\hat{\lambda}^i =  \frac{ \mathcal{I}_t^i}{ {\textstyle \sum_{j=1}^{M}} \mathcal{I}_t^j} \times \mathcal{G},  \quad (i=1, \dots, M).
\label{eq6}
\end{equation}
In this way, conditions with high fitting difficulty will be assigned large weights, as their entries in the difficulty indexes account for a substantial proportion. This weighting way for condition-fitting losses ensures that challenging conditions (high fitting difficulty) receive more attention,  preventing an excessive focus on easier conditions. Hence, all condition-fitting losses can converge at a relatively balanced level.

\begin{algorithm}[t] 
\setlength{\belowcaptionskip}{-10pt}
    \caption{DB-PINNs to dynamically adjust loss weights.} 
    \label{alg1}
    \begin{algorithmic}[1]
     \STATE Consider a PINN with parameters $\theta$ to output $\hat{u}_\theta(\mathbf{x}, t)$. The total loss function is defined as: 
   $\mathcal{L}^{total} = \mathcal{L}^r  + \sum_{i=1}^{M} \lambda^i \mathcal{L}^i$, where
   $\mathcal{L}^r$ denotes the PDE residual loss,  $\mathcal{L}^i$ is $i$-th condition-fitting loss (e.g., the loss term of initial or boundary conditions, observational data, etc.), and $\lambda^i$ is the corresponding loss weight of $\mathcal{L}^i$. 
    \STATE Define $\mathcal{L}_0 = [\mathcal{L}^{1},\dots, \mathcal{L}^{M}]^T \in \mathbb{R}_+^M$: a condition-fitting loss vector at time step $t=0$.
    \STATE Initialize the adaptive weights $\lambda^i = 1, \ (i=1, \dots, M)$;  difficulty indexes $\mathcal{I}_0=[1,\dots,1]^T\in \mathbb{R}_+^M $ and $\mu_{\mathcal{L}_{0}}=\mathcal{L}_{0}$ at time $t=0$. 
        \FOR{$t=1$ to $max\_train\_steps$}
           \STATE  Calculate $\mathcal{L}^r$ and $\mathcal{L}^i$ using respective training points and obtain $\mathcal{L}_t = [\mathcal{L}^{1},\dots, \mathcal{L}^{M}]^T$.
           \STATE  Calculate total gradient ratio: $ \mathcal{G}  = \sum_{i=1}^{M} \frac{\mathrm{max}\left \{ \left | \nabla_\theta \mathcal{L}^r   \right |  \right \} }{\overline{\left | \nabla_\theta \lambda^i  \mathcal{L}^i \right | } }$. \qquad    (\textbf{\textit{Inter-balancing}}) 
             \STATE Update $\mu_{\mathcal{L}_{t}} = (1- \frac{1}{t}) \mu_{\mathcal{L}_{t-1}} + \frac{1}{t}\mathcal{L}_t$. 
            \STATE Calculate $\mathcal{I}_t = \frac{\mathcal{L}_t }{\mu_{\mathcal{L}_{t}} }$.
            \STATE Calculate  $\hat{\lambda}^i =  \frac{ \mathcal{I}_t^i}{ {\textstyle \sum_{j=1}^{M}} \mathcal{I}_t^j} \times \mathcal{G}$.  \qquad      (\textbf{\textit{Intra-balancing}}) 
            \STATE Update the weight $\lambda^i = (1- \frac{1}{t} ) \lambda^i + \frac{1}{t} \hat{\lambda}^i $.   
             \STATE Update the parameters $\theta =  \theta - \eta \nabla_\theta \mathcal{L}^r - \eta \sum_{i=1}^{M}  \lambda^i\nabla_{\theta} \mathcal{L}^i$. 
       \ENDFOR
    \end{algorithmic}
\end{algorithm}

\paragraph{Inner-Balancing for Solving PDEs with One Type of Condition}
Intra-balancing is absent for PDEs with a single specific type of condition (e.g., time-independent PDEs with only BCs). Hence, we introduce inner-balancing instead of intra-balancing to solve those PDEs with a type of condition. Specifically, we decompose the single-condition loss into two or multiple loss terms based on their inner characteristics (e.g., spatial differences, different variables, etc.):  $\mathcal{L}^{total} = \mathcal{L}^r + \lambda\mathcal{L}^{bc}\overset{dec.}{=} \mathcal{L}^r + \sum_i \lambda^i \mathcal{L}^{i}$.
The similar operations in equations (\ref{eq5}) and (\ref{eq6}) can be smoothly carried out to compute weights for each decomposed loss. As a result, inner-balancing enables DB-PINNs to extend beyond the scope of solving time-dependent PDEs.

\subsection{Weight Update Strategy for DB-PINNs} 
Due to the stochastic nature of gradient descent updates, the instant weight values exhibit high variance. Existing gradient-based weighting methods usually adopt the Exponential Moving Average (EMA) strategy to update weights:  $\lambda^i = (1-\alpha ) \lambda^i + \alpha\hat{\lambda}^i$, where $\alpha$ is a hyperparameter. 
However, we observed that this update strategy still fails to handle large variances, leading to frequent occurrences of abrupt spikes and even arithmetic overflow in the instant weight values, particularly when standard deviation or kurtosis is used as a statistical metric. 
To address this issue, we propose a robust free-of-tuning-hyperparameter loss weight update strategy. This strategy uses Welford's algorithm \cite{welford1962note}, an online estimate, to track the mean of the observed condition-fitting loss vector using the following update rules:
\begin{equation} 
\mu_{\mathcal{L}_{t}} = (1- \frac{1}{t}) \mu_{\mathcal{L}_{t-1}} + \frac{1}{t}\mathcal{L}_t,
\end{equation}
where $\mu_{\mathcal{L}_{t}}$ is updated using the previous observations $\mu_{\mathcal{L}_{t-1}}$ and the current observation  $\mathcal{L}_t$. Likewise, the weight is also updated using the same update rules:
\begin{equation}  
\lambda^i = (1- \frac{1}{t}) \lambda^i + \frac{1}{t}\hat{\lambda}^i. 
\end{equation}
This strategy provides a computationally efficient way to estimate the mean from the history of observed losses and robustly update weights without additional hyperparameters. We summarize our proposed method in Algorithm \ref{alg1}.

\section{Experiment}
\subsection{PDE Benchmarks}
We perform the experiments on solving six PDE benchmarks to evaluate the effectiveness of DB-PINN. Due to the page limit, here we report the results of three PDEs. More experimental results are presented in the supplementary material.

\paragraph{Klein-Gordon Equation} 
Klein-Gordon equation is a fundamental nonlinear equation in quantum mechanics and quantum field theory \cite{vemuri2023gradient}. 
We consider the Klein-Gordon equation with a boundary condition $u(x,t) = x\, {\rm cos}(5\pi t) + (xt)^3$ and initial conditions $u(x,0) = x$ and $u_t(x, 0) = 0$: 
\begin{align} 
&u_{tt} - u_{xx} +  u^3 = 0,   & & x \in [0,1] , t\in [0,1]. 
\label{KG_eqn}
\end{align}
Its exact solutions have the same analytical function as the boundary condition.

\paragraph{Wave Equation}
We consider the one-dimensional (1D) wave equation, which conventional PINNs struggle to solve due to its stiffness \cite{wang2022and,mcclenny2020self}.  The wave equation with a boundary condition  $u(x, t) =  0$ as well as the initial conditions $u(x, 0) = {\rm sin}(\pi x) + 0.5\, {\rm sin}(4\pi x)$ and $u_t(x, 0) = 0$ is formulated as: 
\begin{align}
&u_{tt}(x, t) - 4u_{xx}(x, t) = 0, \quad x\in[0, 1], t\in[0,1].
\label{Wave_eqn}
\end{align}
Its exact solutions are given:
\begin{equation}
u(x,t) = {\rm sin}(\pi x)\, {\rm cos}(2\pi t) + 0.5\, {\rm sin}(4\pi x)\, {\rm cos}(8\pi t). 
\label{Wave_exact}
\end{equation}

\paragraph{Helmholtz Equation}
Helmholtz equation describes the propagation of waves with a specific frequency in acoustics, electromagnetics, and fluid dynamics fields. The PDE with a boundary condition $u(x,y) = 0$ is described as follows:
\begin{align} 
&u_{xx} + u_{yy} + u = q(x,y), \  x\in[-1, 1], y\in[-1,1], 
\label{Hel_eqn}
\end{align}
where the forcing term is
\begin{equation}
\begin{aligned}
q(x, y) &= -\pi^2{\rm sin}(\pi x)\,{\rm sin}(4\pi y) -(4\pi)^2{\rm sin}(\pi x)\,{\rm sin}(4\pi y)  \\
& + {\rm sin}(\pi x)\,{\rm sin}(4\pi y),
\label{Hel_eq1}
\end{aligned}
\end{equation}
from which an analytical solution can be derived:
\begin{equation}
u(x, y) = {\rm sin}(\pi x)\,{\rm sin}(4\pi y).
\label{Hel_exact}
\end{equation}
There is no initial condition, and thus no IC loss. To validate the effectiveness of DB-PINN on solving such a time-independent PDE, we partition the four boundaries into two groups: the $x$-axis boundary and the $y$-axis boundary, and thus the BC loss consists of two loss terms. Inner-balancing between these two boundary loss terms aims to achieve a more balanced training for fitting boundary conditions of distinct variables.

\begin{table*}[t] \small 
\centering
\setlength{\tabcolsep}{2.2mm}{
\begin{tabular}{cc|cc|cc|cc}
\bottomrule
    \multicolumn{2}{c|}{\multirow{2}{*}{Method}}             &\multicolumn{2}{c|}{Klein-Gordon Equation} &\multicolumn{2}{c|}{Wave Equation}  &\multicolumn{2}{c}{Helmholtz Equation}\\   \cline{3-8}
    \multicolumn{2}{c|}{}  &L2RE (\%)   &MAE (\%)   &L2RE (\%)  &MAE (\%) &L2RE (\%)   &MAE (\%) \\
\hline
\multicolumn{2}{c|}{Equal weighting}    & 7.272 $\pm$ 3.566 & 2.415 $\pm$ 1.320  &39.870 $\pm$ 1.612 & 16.071 $\pm$ 0.671 & 7.174 $\pm$ 2.434 & 2.257 $\pm$ 0.757  \\
\hline
\multicolumn{2}{c|}{Uncertainty weighting}   & 1.450 $\pm$ 0.705   &0.478 $\pm$ 0.227  & 32.572 $\pm$ 3.486  & 13.094 $\pm$ 1.386 & 0.845 $\pm$ 0.888   & 0.314 $\pm$ 0.368\\ 
\multicolumn{2}{c|}{SA-PINN} & 3.696 $\pm$ 1.085  & 1.193 $\pm$ 0.318  & 33.867 $\pm$ 2.085    & 13.568 $\pm$ 0.826  & 2.413 $\pm$ 0.418  & 0.753 $\pm$ 0.159  \\ \hline
\multicolumn{1}{c|}{\multirow{3}{*}{GW-PINN}}     & mean & 2.062 $\pm$ 0.890   & 0.690 $\pm$  0.293 & 2.126 $\pm$ 0.360   & 0.916 $\pm$ 0.154 & 0.796 $\pm$ 0.125   & 0.188 $\pm$  0.026  \\ 
\multicolumn{1}{c|}{}   & std  & 1.848 $\pm$ 1.455  &0.623 $\pm$  0.494   & \; 8.093 $\pm$ 10.072   & 3.369 $\pm$ 4.117   & 0.270 $\pm$ 0.089  & 0.105 $\pm$ 0.037 \\
\multicolumn{1}{c|}{}   & kurtosis &1.262  $\pm$ 0.403 &0.421  $\pm$ 0.134  & 17.361 $\pm$ 7.938   & 7.630 $\pm$ 3.434  &0.331  $\pm$ 0.071 &0.113  $\pm$ 0.034 \\  \hline
\multicolumn{1}{c|}{\multirow{3}{*}{\makecell[c]{\textbf{DB-PINN} }}}  
& mean & 0.885  $\pm$ 0.257   & 0.297 $\pm$ 0.092 &  0.852 $\pm$ 0.193   & 0.373 $\pm$ 0.084  & 0.180 $\pm$ 0.023   & 0.055 $\pm$ 0.008 \\
\multicolumn{1}{c|}{}   & std  & 0.767 $\pm$ 0.112 &0.254 $\pm$ 0.041  & \textbf{0.292 $\pm$ 0.129}     & \textbf{0.128 $\pm$ 0.056}  & 0.235 $\pm$ 0.065  &0.093 $\pm$ 0.025   \\
\multicolumn{1}{c|}{}   & kurtosis   & \textbf{0.636 $\pm$ 0.132}  &\textbf{0.214 $\pm$ 0.049}  & 0.715 $\pm$ 0.507 & 0.293 $\pm$ 0.214 &\textbf{0.140 $\pm$ 0.018}  &\textbf{0.052 $\pm$ 0.009}  \\  
\toprule
\end{tabular}}
\setlength{\abovecaptionskip}{3pt}
\setlength{\belowcaptionskip}{-5pt}
\caption{Errors (mean $\pm$ std) of different weighting methods for solving these three PDE benchmarks. }  
\label{Table2}
\end{table*}

\subsection{Experimental Setting}
We compare the performance of different weighting methods in PINNs. Specifically, equal weighting is used as a basic baseline.  Uncertainty weighting \cite{xiang2022self} and SA-PINN \cite{mcclenny2023self} are representative of the learning approach in weighting techniques. Regarding the PINNs using gradient-based weighting methods (denoted as GW-PINN), we adopt three different statistical measures which have been utilized in previous works: mean magnitude (mean) \cite{wang2021understanding}, standard deviation (std) \cite{maddu2022inverse}, and kurtosis  \cite{vemuri2023gradient} for depicting the characteristics of gradient distribution. Accordingly, DB-PINN also has three implementations, each corresponding to a distinct gradient statistic used in inter-balancing.

We use hyperbolic tangent activation functions and the Adam optimizer with a learning rate of 0.001 as default.  We choose the $L^2$ relative error (L2RE) and mean absolute error (MAE) as evaluation metrics. In all cases below, the training process is iterated 10 times with random restarts. The average and standard deviation of errors are reported. More detailed experimental setups are shown in the supplementary material.

\subsection{Main Results}

\begin{figure}[!ht] 
  \centering
  \includegraphics[width=\linewidth]{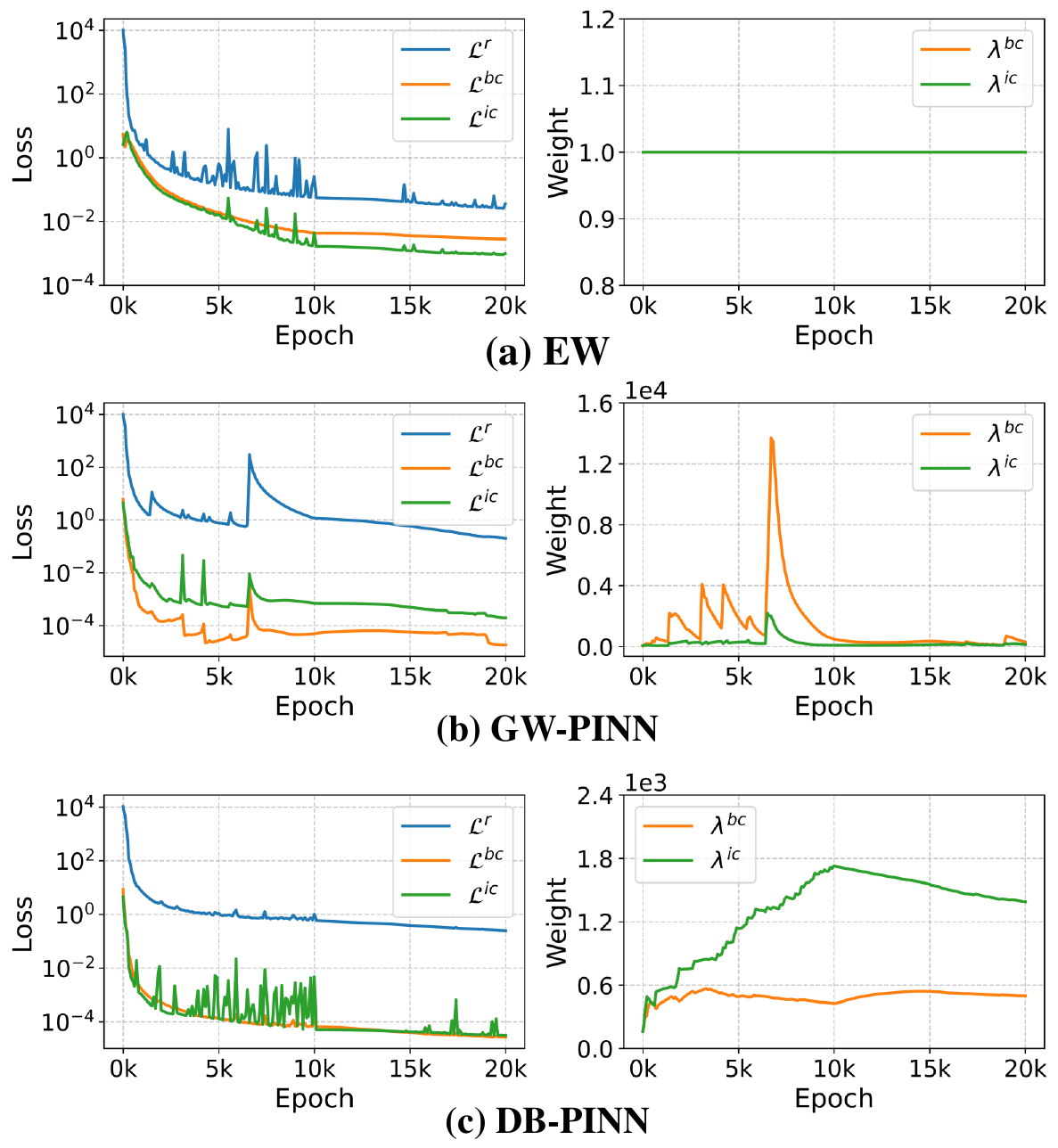}
\setlength{\abovecaptionskip}{-5pt}
\setlength{\belowcaptionskip}{-8pt}
  \caption{ Loss and weight curves during the training on solving the Klein-Gordon equation. 
  }
  \label{KG_loss_weight}
\end{figure}

\paragraph{Insights Behind the Proposed DB-PINN and Its Benefits}  
Figure \ref{KG_loss_weight} illustrates the training loss and weight curves over training epochs for EW, GW-PINN (std), and DB-PINN (std), revealing the limitations of GW-PINN and highlighting the advantages of DB-PINN. First, compared to EW, GW-PINN enlarges the gap between $\mathcal{L}^{bc}$ and $\mathcal{L}^{ic}$. This is because GW-PINN consistently assigns a larger weight to $\mathcal{L}^{bc}$ than  $\mathcal{L}^{ic}$, i.e., the curve of $\lambda^{bc}$ is always above $\lambda^{ic}$, even though the model already fits BC better than IC. Second, training losses usually exhibit variances, manifested as sharp fluctuations in the loss curves in Figure \ref{KG_loss_weight}(a). However, GW-PINN is overly sensitive to instantaneous loss values, and the resulting weights undergo rapid volatility, leading to abrupt spikes. Ultimately, the model fails to reach the optimal. In contrast, DB-PINN effectively mitigates these two issues. In Figure \ref{KG_loss_weight}(c), DB-PINN avoids BC overfitting and IC underfitting, and it ensures both condition losses steadily converge at almost the same rate to values below 10e-4. This is attributed to DB-PINN's capability to dynamically allocate robust weights guided by fitting difficulty with the proposed weight updating strategy. Please refer to the supplementary material for more results.

\begin{figure}[!t] 
  \centering
  \includegraphics[width=1\linewidth]{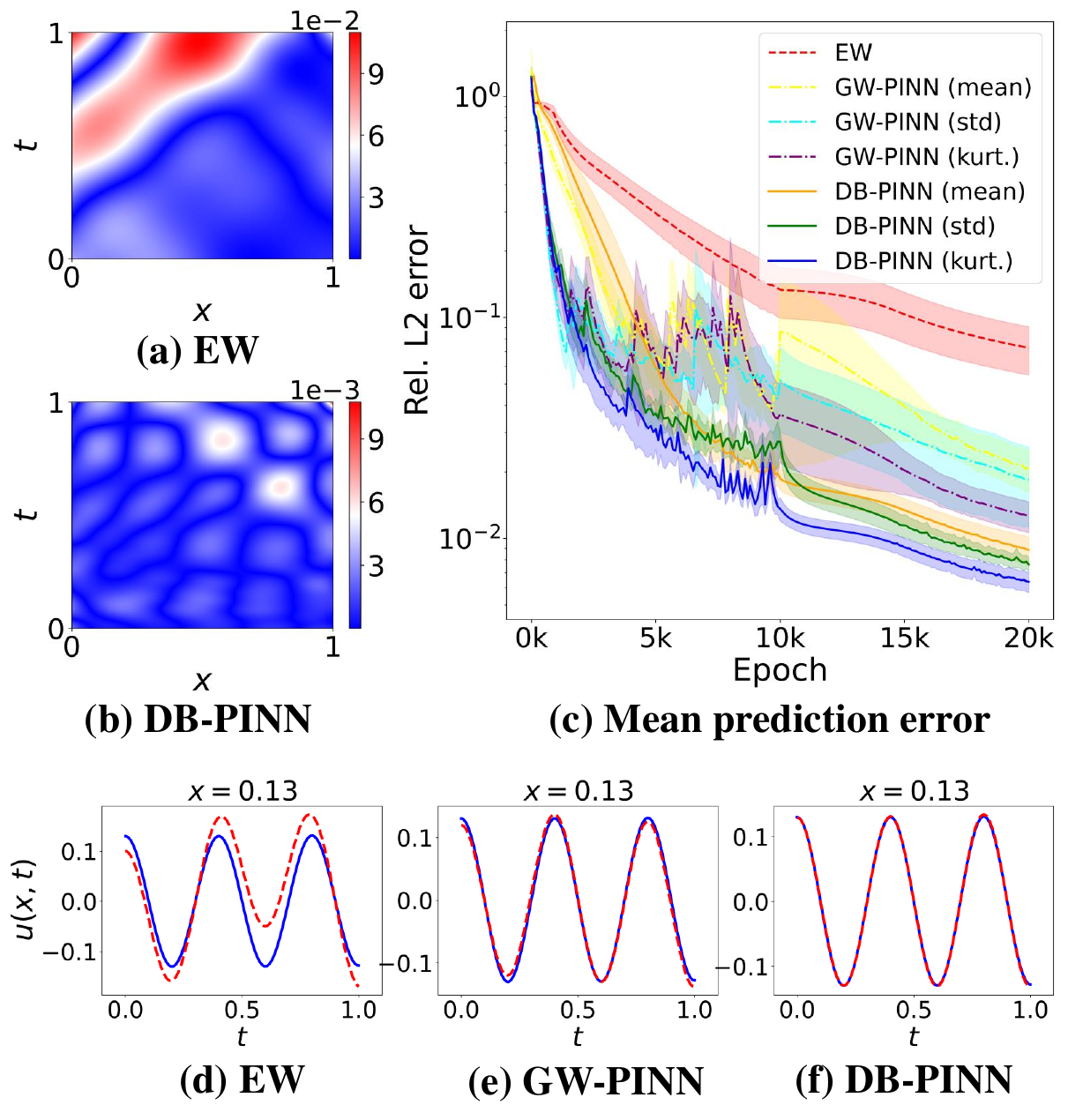}
  \setlength{\abovecaptionskip}{-5pt}
\setlength{\belowcaptionskip}{-3pt}
  \caption{The Klein-Gordon equation. (a-b) Point-wise absolute errors. (c) Mean prediction error curves. (d-f) Comparison between reference solutions (blue) and predictions (red).
  }
  \label{fig1_KG}
\end{figure}

\paragraph{Klein-Gordon Equation} Table \ref{Table2} summarizes the results of PINNs using different weighting methods for solving the Klein-Gordon equation. Compared to equal weighting (EW), which has the largest errors, other methods significantly reduce prediction errors. This proves the importance of finding the optimal weights for better balancing several loss components in the PINN optimization. Regardless of the gradient statistics employed, our proposed DB-PINN consistently achieves more stable and accurate predictions than its counterpart, GW-PINN. DB-PINN (kurtosis) achieves the least prediction errors.  Figure \ref{fig1_KG}(a-b) shows the point-wise absolute prediction errors of EW and DB-PINN (kurtosis), and it indicates that DB-PINN yields the lower maximum errors. The snapshots in Figure \ref{fig1_KG}(d-f) prove again that DB-PINN achieves the best prediction performance. Figure \ref{fig1_KG}(c) shows that DB-PINN converges fastest and eventually achieves the least predictive errors.

\begin{figure}[!t] 
  \centering
  \includegraphics[width=\linewidth]{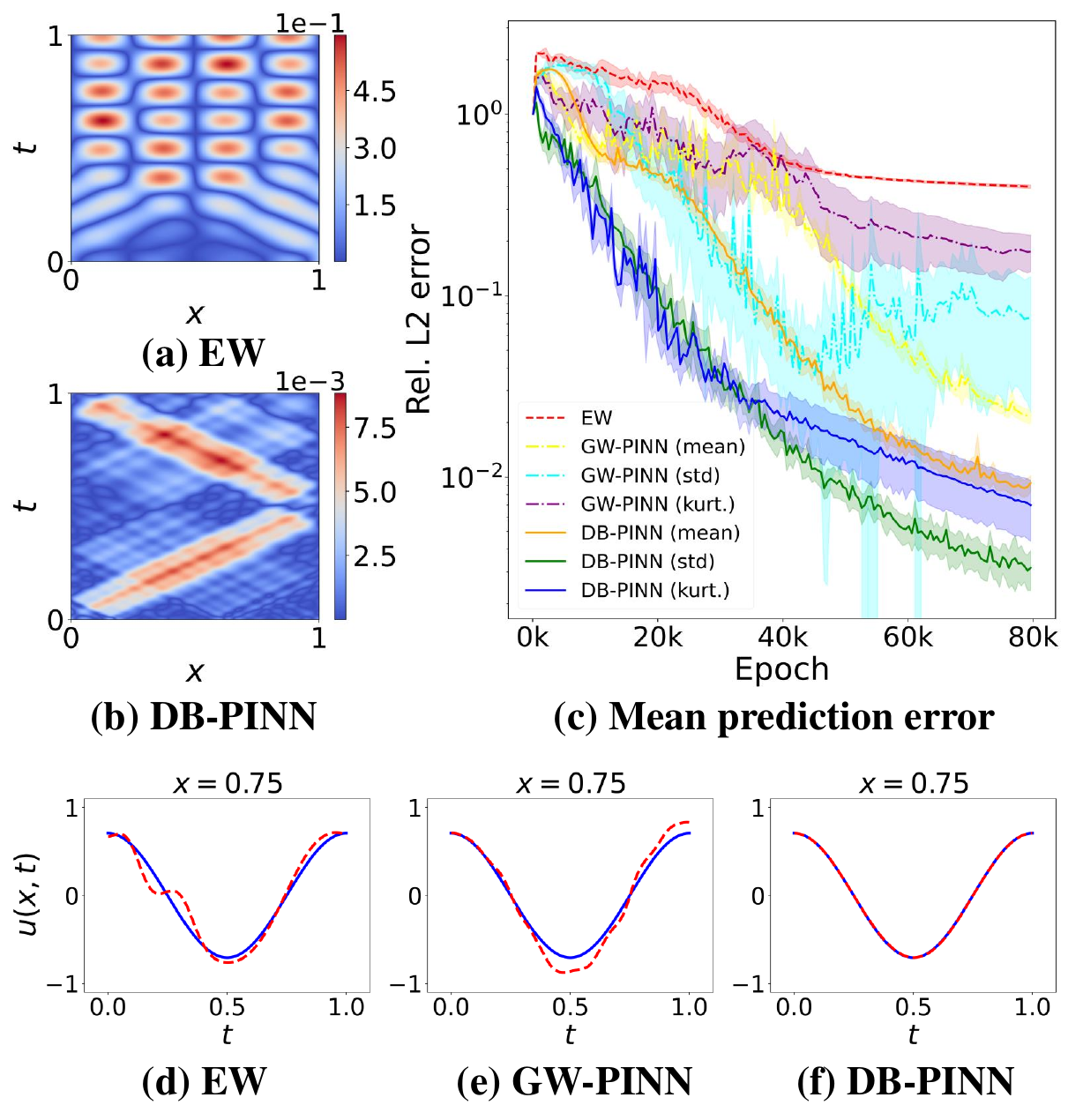}
  \setlength{\abovecaptionskip}{-5pt}
\setlength{\belowcaptionskip}{-3pt}
  \caption{The Wave equation. (a-b) Point-wise absolute errors. (c) Mean prediction error curves. (d-f) Comparison between reference solutions (blue) and predictions (red).
  }
  \label{fig2_Wave}
\end{figure}

\paragraph{Wave Equation} Table \ref{Table2} compares the performance of different methods for solving the wave equation. Equal weighting struggles to solve this PDE with the relative $L^2$ error around 40\%.  Uncertainty weighting and SA-PINN achieve a slight improvement in accuracy. GW-PINN enlarges the performance gap with the baseline but exhibits unstable performance related to the statistical metrics. Thus, the choice of gradient statistics plays a significant role in the performance of GW-PINN when solving more intricate stiff PDEs. Fortunately,  DB-PINN exhibits a significant advantage in that the performance gaps induced by the employed gradient statistics are greatly reduced. Specifically, DB-PINN consistently obtains the best prediction performance among these different weighting methods. Figure \ref{fig2_Wave}(a-b) visually compares the performance of EW and DB-PINN (kurtosis). DB-PINN achieves better prediction accuracy, which is also proved by the comparison in Figure \ref{fig2_Wave}(d-f). Figure \ref{fig2_Wave}(c) shows that DB-PINN has a fast convergence speed and stable predictive performance with low errors.

\begin{figure}[!ht] 
  \centering
  \includegraphics[width=\linewidth]{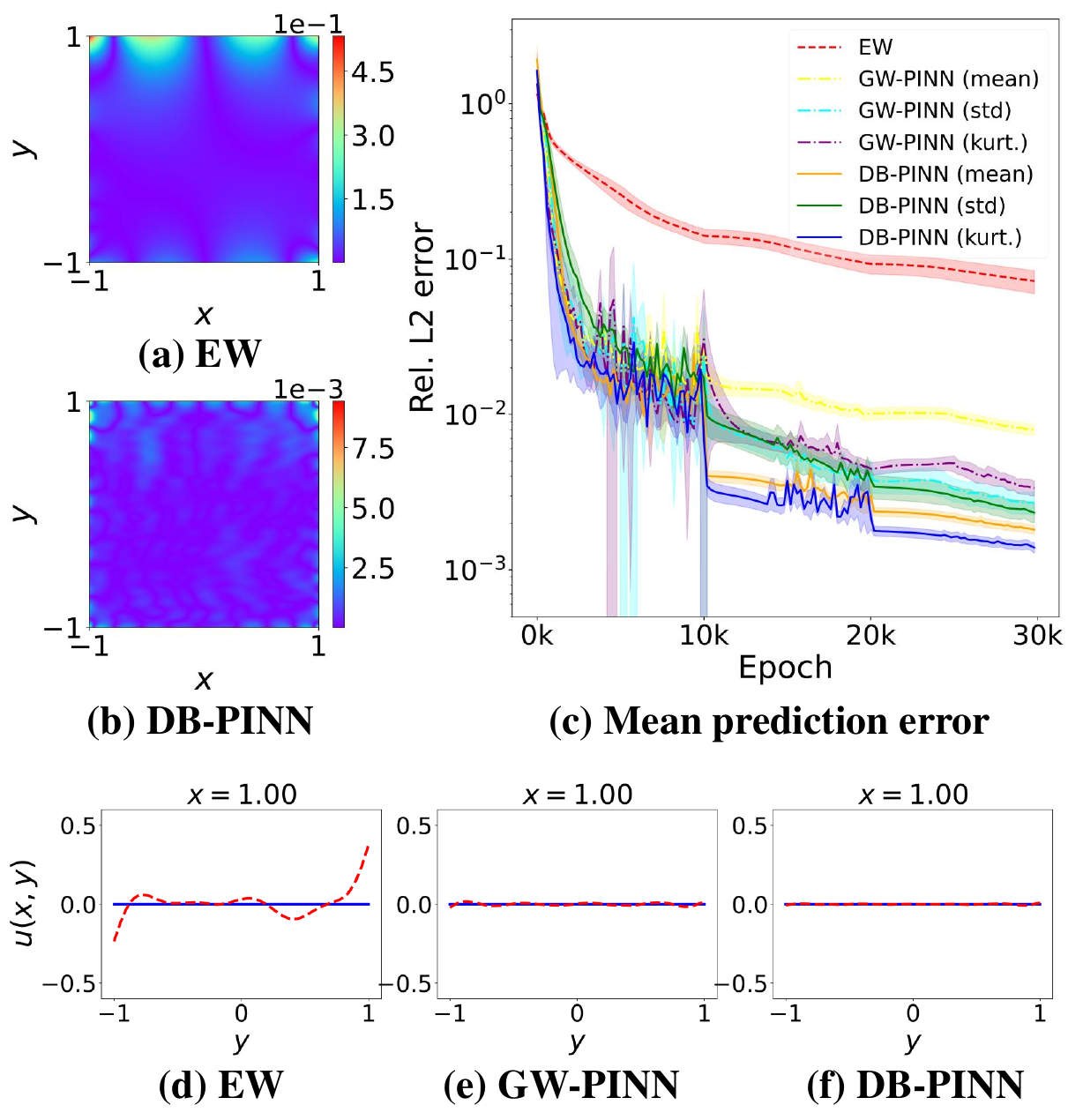}
  \setlength{\abovecaptionskip}{-5pt}
\setlength{\belowcaptionskip}{-5pt}
  \caption{The Helmholtz equation. (a-b) Point-wise absolute errors. (c) Mean prediction error curves. (d-f) Comparison between reference solutions (blue) and predictions (red).
  }
  \label{fig3_Hel}
\end{figure}

\paragraph{Helmholtz Equation} Table \ref{Table2} reports the results for solving the Helmholtz PDE. Compared with GW-PINN using identical statistics, DB-PINN attains lower errors. DB-PINN (kurtosis) yields the least errors with the L2RE of 0.140\% and MAE of 0.052\%. Hence, the effectiveness of DB-PINN in solving time-independent PDEs also has been validated. Figure \ref{fig3_Hel}(a-b) indicates that DB-PINN (kurtosis) has the lowest maximum errors. Figure \ref{fig3_Hel}(d-f) shows the snapshots of prediction at the boundary $x=1.00$. DB-PINN (kurtosis) achieves predictions that closely match reference solutions. Figure \ref{fig3_Hel}(c) shows that DB-PINN exceeds GW-PINN in both convergence speed and prediction accuracy.

\subsection{Ablation Study}
\paragraph{Dual-Balancing} We conduct ablation studies on solving the Klein-Gordon equation to evaluate the respective contribution of the dual-balancing and the weight update strategy in DB-PINN. Table \ref{Table4} reports the results of the ablation study about the dual-balancing mechanism in DB-PINN. Our approach without inter- and intra-balancing is denoted as `DB-PINN w/o balancing', which is equivalent to GW-PINN with the proposed weight update strategy. `DB-PINN-Avg' refers to the DB-PINN without intra-balancing, which allocates the average of the aggregated weight to each condition loss. Regardless of any one of the statistical metrics, `DB-PINN w/o balancing' consistently achieves inferior performance than DB-PINN in Table \ref{Table2}. Notice that `DB-PINN-Avg' obtains poorer accuracy than DB-PINN, which justifies the necessity of balancing the training to learn multiple conditions with different levels of fitting difficulty. Therefore, inter-balancing and intra-balancing are indispensable to addressing two imbalance issues, constituting the dual-balancing mechanism collaboratively.

\begin{table}[t] \small 
\centering
\setlength{\tabcolsep}{2.5mm}{ 
\begin{tabular}{cc|c|c}
\bottomrule
\multicolumn{1}{c|}{Method} &\multicolumn{1}{c|}{Stats}    &\multicolumn{1}{c|}{L2RE (\%)} & \multicolumn{1}{c}{MAE (\%)} \\ 
\hline
\multicolumn{1}{c|}{\multirow{3}{*}{\makecell[c]{DB-PINN \\  w/o balancing}}}     & mean & 1.095 $\pm$ 0.142  & 0.376 $\pm$ 0.049 \\
\multicolumn{1}{c|}{}   & std  & 0.969 $\pm$ 0.204  &0.329 $\pm$ 0.070 \\
\multicolumn{1}{c|}{}   & kurt.  &0.797 $\pm$ 0.122  &0.267 $\pm$ 0.043 \\  \hline
\multicolumn{1}{c|}{\multirow{3}{*}{\makecell[c]{DB-PINN-Avg }}}     & mean & 0.985  $\pm$ 0.175   & 0.336 $\pm$ 0.062 \\
\multicolumn{1}{c|}{}   & std   & 0.899 $\pm$ 0.152  &0.302 $\pm$ 0.052  \\
\multicolumn{1}{c|}{}   & kurt.   &0.776 $\pm$ 0.113  &0.263 $\pm$ 0.038 \\ 
\toprule
\end{tabular}}
   \setlength{\abovecaptionskip}{1pt}
\setlength{\belowcaptionskip}{-4pt}
\caption{Ablation study of the dual-balancing mechanism in DB-PINN (w/o denotes without). }
\label{Table4}
\end{table}

\begin{table}[t] \small 
\centering
\setlength{\tabcolsep}{2mm}{ 
\begin{tabular}{ccc|c|c}
\bottomrule
\multicolumn{1}{c|}{Method}  &\multicolumn{1}{c|}{$\alpha$}    &\multicolumn{1}{c|}{Stats}           & \multicolumn{1}{c|}{L2RE (\%)} & \multicolumn{1}{c}{MAE (\%)} \\ 
\hline
\multicolumn{1}{c|}{\multirow{9}{*}{\makecell[c]{DB-PINN \\ w/ EMA }}} &\multicolumn{1}{c|}{\multirow{3}{*}{0.5}} & mean & 1.486  $\pm$ 0.534   & 0.490 $\pm$ 0.163 \\
\multicolumn{1}{c|}{}  &\multicolumn{1}{c|}{} & std   & 1.337 $\pm$ 0.367  &0.451 $\pm$ 0.124 \\
\multicolumn{1}{c|}{}  &\multicolumn{1}{c|}{} & kurt.   & 1.205 $\pm$ 0.273  &0.405 $\pm$ 0.092 \\  \cline{2-5}
\multicolumn{1}{c|}{}  &\multicolumn{1}{c|}{\multirow{3}{*}{0.2}}   & mean &  1.305 $\pm$ 0.474    & 0.440 $\pm$ 0.164 \\
\multicolumn{1}{c|}{}  &\multicolumn{1}{c|}{}  & std   & 1.182 $\pm$ 0.294  &0.402 $\pm$ 0.098   \\
\multicolumn{1}{c|}{}   &\multicolumn{1}{c|}{} & kurt.   & 1.112 $\pm$ 0.248   & 0.378 $\pm$ 0.086 \\  \cline{2-5}
\multicolumn{1}{c|}{}  &\multicolumn{1}{c|}{\multirow{3}{*}{0.01}}   & mean &  1.034 $\pm$  0.306  &  0.347 $\pm$ 0.102 \\
\multicolumn{1}{c|}{}   &\multicolumn{1}{c|}{} & std   & 0.913 $\pm$ 0.133  & 0.307 $\pm$ 0.046  \\
\multicolumn{1}{c|}{}   &\multicolumn{1}{c|}{} & kurt.   & 0.859 $\pm$ 0.193  & 0.292 $\pm$ 0.067 \\  
\toprule
\end{tabular}}
   \setlength{\abovecaptionskip}{1pt}
 \setlength{\belowcaptionskip}{-3pt}
\caption{Ablation study of the weight update strategy in DB-PINN (w/ EMA denotes that the update strategy is replaced with EMA). } 
\label{Table5}
\end{table}

\paragraph{Weight Update Strategy} 
To evaluate the proposed weight update strategy, we replace it with EMA,  denoted as `DB-PINN w/ EMA'. The results are reported in Table \ref{Table5}. As there is a hyperparameter $\alpha$ in EMA, we run `DB-PINN w/ EMA' with $\alpha=0.5$, $\alpha=0.2$, and  $\alpha=0.01$, respectively. Table \ref{Table5} indicates that the performance of `DB-PINN w/ EMA' is very susceptible to $\alpha$. The accuracy of `DB-PINN w/ EMA' increases along with the descending $\alpha$. This reminds us to carefully select the value of $\alpha$ when using EMA, but undoubtedly, it can lead to the heavy consumption of computational resources because of trials and errors in tuning this hyperparameter. Importantly, `DB-PINN w/ EMA' steadily performs worse than DB-PINN, which demonstrates the superiority of the weight update strategy over EMA.

\section{Conclusion}
In this work, we propose DB-PINNs to adaptively adjust the loss weights in PINNs by combining inter-balancing (to mitigate the gradient imbalance between PDE residual loss and condition losses) with intra-balancing (to address the imbalance in fitting difficulty among different condition losses). We further introduce a robust weight update strategy to tackle the large variance in weight values for stable training. Experimental results show that DB-PINN consistently exceeds GW-PINN in both convergence speed and prediction accuracy.

\bibliographystyle{named}
\bibliography{ijcai25}

\end{document}